\documentstyle[12pt,chicagob]{article}

\newtheorem{THEOREM}{Theorem}[section]
\newenvironment{theorem}{\begin{THEOREM} \hspace{-.85em} {\bf :} }%
                        {\end{THEOREM}}
\newtheorem{LEMMA}[THEOREM]{Lemma}
\newenvironment{lemma}{\begin{LEMMA} \hspace{-.85em} {\bf :} }%
                      {\end{LEMMA}}
\newtheorem{COROLLARY}[THEOREM]{Corollary}
\newenvironment{corollary}{\begin{COROLLARY} \hspace{-.85em} {\bf :} }%
                          {\end{COROLLARY}}
\newtheorem{PROPOSITION}[THEOREM]{Proposition}
\newenvironment{proposition}{\begin{PROPOSITION} \hspace{-.85em} {\bf :} }%
                            {\end{PROPOSITION}}
\newtheorem{DEFINITION}[THEOREM]{Definition}
\newenvironment{definition}{\begin{DEFINITION} \hspace{-.85em} {\bf :} \rm}%
                            {\end{DEFINITION}}
\newtheorem{CLAIM}[THEOREM]{Claim}
\newenvironment{claim}{\begin{CLAIM} \hspace{-.85em} {\bf :} \rm}%
                            {\end{CLAIM}}
\newtheorem{EXAMPLE}[THEOREM]{Example}
\newenvironment{example}{\begin{EXAMPLE} \hspace{-.85em} {\bf :} \rm}%
                            {\end{EXAMPLE}}
\newtheorem{REMARK}[THEOREM]{Remark}
\newenvironment{remark}{\begin{REMARK} \hspace{-.85em} {\bf :} \rm}%
                            {\end{REMARK}}

\newcommand{\thm}{\begin{theorem}}
\newcommand{\lem}{\begin{lemma}}
\newcommand{\pro}{\begin{proposition}}
\newcommand{\dfn}{\begin{definition}}
\newcommand{\rem}{\begin{remark}}
\newcommand{\xam}{\begin{example}}
\newcommand{\cor}{\begin{corollary}}

\newcommand{\ethm}{\end{theorem}}
\newcommand{\elem}{\end{lemma}}
\newcommand{\epro}{\end{proposition}}
\newcommand{\edfn}{\bbox\end{definition}}
\newcommand{\erem}{\bbox\end{remark}}
\newcommand{\exam}{\bbox\end{example}}
\newcommand{\ecor}{\end{corollary}}
\newcommand{\eprf}{\bbox\vspace{0.1in}}
\newcommand{\beqn}{\begin{equation}}
\newcommand{\eeqn}{\end{equation}}

\newcommand{\bbox}{\vrule height7pt width4pt depth1pt}

\newcommand{\clm}{\begin{claim}}
\newcommand{\eclm}{\end{claim}}

\newcommand{\sat}{\models}

\newcommand{\rimp}{\Rightarrow}

\newcommand{\union}{\cup}

\renewcommand{\phi}{\varphi}

\newcommand{\B}{{\cal B}}

\newcommand{\F}{{\cal F}}

\newcommand{\K}{{\cal K}}
\newcommand{\M}{{\cal M}}

\newcommand{\R}{{\cal R}}

\newcommand{\U}{{\cal U}}
\newcommand{\V}{{\cal V}}

\newcommand{\<}{\langle}
\renewcommand{\>}{\rangle}

\newcommand{\ie}{i.e.,~}

\newcommand{\ol}{\setlength{\itemsep}{0pt}\begin{enumerate}}
\newcommand{\eol}{\end{enumerate}\setlength{\itemsep}{-\parsep}}
\newcommand{\ul}{\setlength{\itemsep}{0pt}\begin{itemize}}
\newcommand{\dl}{\setlength{\itemsep}{0pt}\begin{description}}
\newcommand{\edl}{\end{description}\setlength{\itemsep}{-\parsep}}
\newcommand{\eul}{\end{itemize}\setlength{\itemsep}{-\parsep}}

\newcommand{\ES}{E_\cS}

\newcommand{\true}{\mbox{{\it true}}}

\newcommand{\commentout}[1]{}

\newcommand{\bi}{\begin{itemize}}
\newcommand{\ei}{\end{itemize}}
\newcommand{\be}{\begin{enumerate}}
\newcommand{\ee}{\end{enumerate}}

\newcommand{\PrB}{\Pr}
\renewcommand{\S}{{\cal S}}
\newcommand{\ML}{\mbox{{\it ML}}}

\newcommand{\FB}{\mbox{{\it FB}}}
\newcommand{\AS}{\mbox{{\it AS}}}
\renewcommand{\ES}{\mbox{{\it ES}}}

\begin{document}

\begin{titlepage}
\title{Causes and Explanations: A
Structural-Model
Approach.
Part II: Explanations}
\author{Joseph Y. Halpern%
\thanks{Supported in part by NSF under
grants IRI-96-25901 and 0090145.}\\
Cornell University\\
Dept. of Computer Science\\
Ithaca, NY 14853\\
halpern@cs.cornell.edu\\
http://www.cs.cornell.edu/home/halpern\\
\and
Judea Pearl%
\thanks{Supported in part by grants from NSF,
ONR, AFOSR, and MICRO.}\\
Dept. of Computer Science\\
University of California, Los Angeles\\
Los Angeles, CA 90095\\
judea@cs.ucla.edu\\
http://www.cs.ucla.edu/$\sim$judea
}

\date{\today}
\setcounter{page}{0}
\thispagestyle{empty}
\maketitle
\thispagestyle{empty}

\begin{abstract}
We propose new definitions of 
{\em (causal) explanation},
using
{\em structural equations\/} to model counterfactuals.
The definition is based on the notion of {\em actual
cause\/}, as 
defined and motivated in a companion paper.  
Essentially,
an explanation is a fact
that is not known for certain but, if found to be true,
would constitute an actual cause of the  fact to
be explained, regardless of the agent's initial uncertainty.   
We show that the definition handles well
a number of problematic examples from the literature.
\end{abstract}
\end{titlepage}

\section{Introduction}
The automatic generation of adequate explanations is
a task essential in planning, diagnosis and natural language processing.
A system doing inference must be able to explain its
findings and recommendations to evoke 
a user's confidence.
However, getting a good definition of explanation is a
notoriously difficult problem, which has been studied 
for years.
(See \cite{CH97,Gardenfors1,Hempel65,Pearl,Salmon89} and the references
therein for an introduction to and discussion of the issues.)

In Part I of this paper \cite{HP01b}
we give
a definition of actual causality
using structural equations.  Here we show how the ideas behind that
definition can be used to give an elegant definition of (causal)
explanation that deals well with many of the problematic examples discussed in the
literature.  The basic idea is that an explanation is a fact
that is not known for certain but, if found to be true,
would constitute an actual cause of the {\em explanandum\/} (the fact to
be explained), regardless of the agent's initial uncertainty.   

Note that our definition involves causality and knowledge.
Following G\"ardenfors \citeyear{Gardenfors1}, we take the notion of
explanation to be relative to an agent's epistemic state.  What counts
as an explanation for one agent may not count as an explanation for
another agent.  We also follow G\"ardenfors in allowing explanations to
include (fragments of) a causal model.  To borrow an example from
G\"ardenfors, an agent seeking 
an explanation of why Mr.\ Johansson has been taken ill with lung cancer
will not consider the fact that he worked for years in asbestos
manufacturing a part of an explanation if he already knew this fact.
For such an agent, an explanation of Mr.~Johansson's illness may
include a causal model describing  the connection between asbestos
fibers and lung cancer.  On the other hand, for someone who already
knows the causal model but does not know that Mr.\ Johansson worked in
asbestos manufacturing, the explanation would involve Mr.\ Johansson's
employment but would not mention the causal model.

Where our definition differs from that of G\"ardenfors (and all others in
the literature) is in the way it is formulated in terms of 
the underlying notions of knowledge, causal models, actual
causation, and counterfactuals. The definition is not based on
probabilistic dependence, ``statistical relevance'', or
logical implication, and thus
is able to deal with the directionality inherent in common explanations.
While it seems reasonable to say ``the height of the flag pole explains 
the length of the shadow'', it would sound awkward if one
were to explain the former with the latter.  Our definition is able to
capture this distinction easily.

The best judge of the
adequacy of an approach are the intuitive appeal of the definitions and
how well it deals with examples; we 
believe that this paper shows that our
approach fares well on both counts.

The remainder of the paper is organized as follows.  In
Section~\ref{sec:models}, we review the basic definitions of causal
models based on structural equations, which are the basis for our 
definitions of causality and explanation, and then review the definition
of causality from the companion paper.  We have tried to include enough
detail here to make the paper self-contained, but we encourage the
reader to consult the companion paper for more motivation and discussion.
In Section~\ref{sec:explanation} we give the basic definition of 
explanation, under the assumption that
the causal model is known.
In Section~\ref{s:partialexp}, probability is added to the picture, to
give notions of {\em partial explanation} and {\em explanatory power}.
The general definition, which dispenses with the assumption that the
causal model is known,
is discussed in Section~\ref{sec:general}.
We conclude in Section~\ref{sec:discussion} with some discussion.

\section{Causal Models and the Definition of Actual Causality: A
Review}\label{sec:models} 
To make this paper self-contained, this section repeats material
from the companion paper; we
review the basic definitions of causal models,
as defined in terms of structural equations, 
the syntax and semantics of a language for reasoning about
causality and explanations, and the definition of actual cause.

\subsection{Causal models}

The use of structural equations as a model for causal relationships
is standard in the social sciences,
and seems to go back to the work of Sewall Wright in the 1920s (see
\cite{Goldberger72} for a discussion);
the particular framework that we use here
is due to Pearl \citeyear{Pearl.Biometrika}, and is further developed in
\cite{pearl:2k}.  

The basic picture 
is that the world is described by random variables,
some of which may have a causal
influence on others. This influence
is modeled by a set of {\em structural equations}.
Each equation represents a distinct mechanism (or law) in the
world, which may be modified (by external actions) without
altering the others.
In practice, it
seems useful to split the random variables into two sets, the {\em
exogenous\/} variables, whose values are determined by factors outside
the model, and the {\em endogenous\/} variables,
whose values are ultimately determined by the exogenous variables.
It is these endogenous
variables whose values are described by the structural equations.

Formally, a {\em signature\/} $\S$ is a tuple $(\U,\V,\R)$,
where $\U$ is a 
set of exogenous variables, $\V$ is a finite 
set of endogenous variables,
and $\R$ associates with every variable $Y \in
\U \union \V$ a nonempty set $\R(Y)$ of possible values for $Y$
(that is, the set of values over which $Y$ {\em ranges}).
A {\em causal\/} (or {\em structural\/}) {\em model\/} 
over signature $\S$ is a tuple $M=(\S,\F)$,
where $\F$ associates with each variable $X \in \V$ a function denoted
$F_X$ such that $F_X: (\times_{U \in \U} \R(U))
\times (\times_{Y \in \V - \{X\}} \R(Y)) \rightarrow \R(X)$.
$F_X$ determines the value of $X$
given the values of all the other variables in $\U \union \V$.
For example, if $F_X(Y,Z,U) = Y+U$ (which we usually write 
as $X=Y+U$), then if $Y = 3$ and $U = 2$, then $X=5$,
regardless of how $Z$ is set.

These equations can be thought of as representing 
processes (or mechanisms) by which values are assigned to
variables.  Hence, like physical laws, they support a
counterfactual interpretation.  For example, the equation above
claims that, in the context $U = u$,
if $Y$ were $4$, then $X$ would
be $u+4$ (which we write as $(M,u) \sat 
[Y \gets 4](X=u+4)$), regardless of what values 
$X$, $Y$, and $Z$
actually take in the real world. 

The counterfactual interpretation and the causal
asymmetry associated with the structural equations are best
seen when we consider external interventions (or spontaneous
changes), under which some equations in $F$ are modified. 
An equation such as 
$x = F_X(\vec{u},y)$ should be thought of as saying
that in a context where the exogenous variables have values $\vec{u}$,
if $Y$ were set to $y$ 
by some means (not specified in the model), 
then $X$ would take on the value 
$x$, as dictated by $F_X$. The same does not hold when we intervene
directly on $X$; 
such an intervention amounts to assigning a value to $X$ by
external means, thus overruling the assignment specified
by $F_X$. 

For those more comfortable with thinking of counterfactuals in terms of
possible worlds, this modification of equations  
may be given a simple ``closest world'' interpretation:  
the solution of the equations obtained
replacing the equation for $Y$ with the equation $Y = y$,
while leaving all other equations unaltered,
gives the closest ``world'' to the actual world where $Y=y$.

\commentout{
We typically write $X = F_X(\U,\V-\{X\})$, to show that $X$ is
determined by the setting of the remaining variables.
The functions $F_X$ define a set of {\em (modifiable)
structural equations}, relating the values of the variables.  Because
$F_X$ is a function, there is a unique value of $X$ once we have set all
the other variables.
Notice that we have such functions only for the endogenous variables.
The exogenous variables
are taken as given; it is their effect on the endogenous
variables (and the effect of the endogenous variables on each other)
that we are modeling with the structural equations.
}

We can describe (some salient features of) a causal model $M$ using a
{\em causal network}.  
This is a graph
with nodes corresponding to the random variables in $\V$ and an edge
from a node labeled $X$ to one labeled $Y$ if $F_Y$ depends on the value
of $X$.
Intuitively, variables can have a causal effect only on their
descendants in the causal network; if $Y$ is not a descendant of $X$,
then a change in the value of $X$ has no effect on the value of $Y$.
Causal networks  are similar in spirit to Lewis's
{\em neuron diagrams} \citeyear{Lewis73}, but there are significant
differences as well (see Part I for a discussion).
In this paper, 
we restrict attention to what are called {\em
recursive\/} (or {\em acyclic\/}) equations; 
these are ones that can be
described with a causal network that is a directed acyclic graph (that
is, a graph that has no  cycle of edges).
It should be clear that if $M$ is a recursive causal model,
then there is always a
unique solution to the equations in
$M$, given a setting $\vec{u}$ for the
variables in $\U$.%
\footnote{In the companion paper, there is some discussion on how to
extend the definition of causality to nonrecursive models.  It seems
that these ideas should apply to our definition of explanation as well.
However, we do not pursue this issue here.}
Such a setting is called a {\em context}.  Contexts will play the role
of possible worlds when we model uncertainty.
For future reference, a pair $(M,\vec{u})$ consisting of a causal model
and a context is called a {\em situation}.

\begin{figure}[htb]
\input{psfig}
\centerline{\psfig{figure=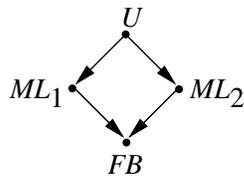}}
\caption{The causal network for $M_1$ and
$M_2$.}
\label{fig1m}
\end{figure}

\xam\label{xam:arson}
Suppose that two arsonists
drop lit matches in different parts of a dry forest, and both
cause trees to start burning.
Consider two scenarios.  In the first, called 
the {\em disjunctive scenario},
either match by itself
suffices to burn down the whole forest.  That is, even if only one
match were lit, the forest would burn down.  In the second scenario,
called the {\em conjunctive scenario}, both matches are necessary to burn down
the forest; if only one match were lit, the fire would die down
before the forest was consumed.
We can describe the essential structure of these two
scenarios using a causal model with four variables:
\begin{itemize}
\item an exogenous variable $U$ that determines,
among other things, the motivation and state of mind of
the arsonists.  For simplicity, assume that $\R(U) = \{u_{00}, u_{10},
u_{01}, u_{11}\}$; if $U=u_{ij}$, then the first arsonist intends to
start a fire iff $i=1$ and the second arsonist intends to start a fire
iff $j=1$.  In both scenarios $U=u_{11}$.
\item endogenous variables $\ML_1$ and $\ML_2$, each either 0 or 1, where
$\ML_i=0$ if arsonist $i$ doesn't drop the match and $\ML_i=1$ if he does,
for $i =1, 2$.
\item an endogenous variable $\FB$ for forest burns down, with values 0
(it doesn't) and 1 (it does).
\end{itemize}
Both scenarios have the same causal network
(see Figure \ref{fig1m});
they differ only in the equation for $\FB$.
For the disjunctive scenario
we have $F_{FB}(u,1,1) = F_{FB}(u,0,1) =
F_{FB}(u,1,0) = 1$ and $F_{FB}(u,0,0) = 0$ (where $u \in \R(U)$);
for the conjunctive scenario
we have $F_{FB}(u,1,1) = 1$ and
$F_{FB}(u,0,0) = F_{FB}(u,1,0) = F_{FB}(u,0,1) = 0$.

In general, we expect that the causal
model for reasoning about forest fires would involve many other
variables; in particular, variables for other potential causes
of forest fires such as lightning and unattended campfires.  Here we focus
on that part of the causal model that involves forest fires started by
arsonists.  Since for causality we assume that all the relevant facts
are given, we can assume here that it is known that there were no
unattended campfires and there was no lightning, which makes it safe to
ignore that portion of the causal model.

Denote by $M_1$ and $M_2$ the
(portion of the)
causal models associated with
the disjunctive and conjunctive scenarios, respectively.  The causal
network for the relevant portion of $M_1$ and $M_2$ is described in
Figure~\ref{fig1m}.
The diagram emphasizes that the value of $FB$ is determined by the
values of $\ML_1$ and $\ML_2$ (which in turn are determined by the
value of the exogenous variable $U$.
\exam

As we said,
a causal model has the resources to determine counterfactual effects.
Given a causal model $M = (\S,\F)$, a (possibly
empty)  vector
$\vec{X}$ of variables in $\V$, and vectors $\vec{x}$ and
$\vec{u}$ of values for the variables in
$\vec{X}$ and $\U$, respectively, we can define a new causal model
denoted
$M_{\vec{X} \gets \vec{x}}$ over the signature $\S_{\vec{X}}
= (\U, \V - \vec{X}, \R|_{\V - \vec{X}})$.
Formally, $M_{\vec{X} \gets \vec{x}} = (\S_{\vec{X}},
\F^{\vec{X} \gets \vec{x}})$,
where $F_Y^{\vec{X} \gets \vec{x}}$ is obtained from $F_Y$
by setting the values of the
variables in $\vec{X}$ to $\vec{x}$.
Intuitively, this is the causal model that results when the variables in
$\vec{X}$ are set to $\vec{x}$ by some external action
that affects only the variables in $\vec{X}$;
we do not model the action or its causes explicitly.
For example, if $M_1$ is the model for the disjunctive scenario in
Example~\ref{xam:arson}, then $(M_1)_{{\it ML}_1 \gets 0}$ is the model
where $\FB = \ML_2$: if the first match is not dropped, then there is a
fire if and only if the second match is dropped.
Similarly,
$(M_2)_{{\it ML}_2 \gets 0}$ is the model where $\FB = 0$: if the first
match is not dropped, then there is no fire in the conjunctive scenario.
Note that if $M$ is a recursive causal model, then there is always a
unique solution to the equations in $M_{\vec{X} \gets \vec{x}}$ for all
$\vec{X}$ and $\vec{x}$.


\subsection{Syntax and Semantics:}
Given a signature $\S = (\U,\V,\R)$, a formula of the form $X = x$, for
$X \in V$ and $x \in \R(X)$, is called a {\em primitive event}.   A {\em
basic causal formula\/} 
is one of the form
$[Y_1 \gets y_1, \ldots, Y_k \gets y_k] \phi$,
where 
\begin{itemize}
\item 
$\phi$ is a Boolean
combination of primitive events;
\item $Y_1,\ldots, Y_k$ are distinct variables in $\V$;  and
\item 
$y_i \in \R(Y_i)$.
\end{itemize}
Such a formula is 
abbreviated
as $[\vec{Y} \gets \vec{y}]\phi$.
The special
case where $k=0$
is abbreviated as
$\phi$.
Intuitively, $[Y_1 \gets y_1, \ldots, Y_k \gets y_k] \phi$ says that
$\phi$ holds in the counterfactual world that would arise if
$Y_i$ were set to $y_i$, $i = 1,\ldots,k$.
A {\em causal formula\/} is a Boolean combination of basic causal
formulas.%
\footnote{
If we write $\rightarrow$ for conditional implication, then
a formula such as $[Y\gets y] \phi$ can be written as $Y = y \rightarrow
\phi$: if $Y$ were $y$, then $\phi$ would hold.  We use the present
notation to emphasize the fact that, although we are viewing $Y \gets y$
as a modal operator, we are not giving semantics using the standard
possible-worlds approach.}

A causal formula $\phi$ is true or false in a causal model, given a context.
We write $(M,\vec{u}) \sat \phi$ if
$\phi$ is true in
causal model $M$ given context $\vec{u}$.
\commentout{
\footnote{We remark that in \cite{GallesPearl97,Hal20},
the context $\vec{u}$ does not appear on the left-hand side of $\sat$;
rather, it is incorporated in the formula $\phi$ on the right-hand (so
that a basic formula becomes $X(\vec{u}) = x$).  
Additionally, Pearl \citeyear{pearl:2k} abbreviated
$(M,\vec{u}) \sat [\vec{Y} \gets \vec{y}](X = x)$
as
$X_{y}(u)=x$. The presentation here
makes certain things more explicit, although they are technically
equivalent.}}
$(M,\vec{u}) \sat [\vec{Y} \gets \vec{y}](X = x)$ if 
the variable $X$ has value $x$ 
in the
unique (since we are dealing with recursive models) solution
to
the equations in
$M_{\vec{Y} \gets \vec{y}}$ in context $\vec{u}$ (that is, the
unique vector
of values for the exogenous variables that simultaneously satisfies all
equations $F^{\vec{Y} \gets \vec{y}}_Z$, $Z \in \V - \vec{Y}$,
with the variables in $\U$ set to $\vec{u}$).
We extend the definition to arbitrary causal formulas
in the obvious way.

Thus, in Example~\ref{xam:arson}, we have
$(M_{1}, u_{11}) \sat [\ML_1 = 0](\FB = 1)$ and 
$(\M_2, u_{11}) \sat [\ML_1 = 0](\FB = 0)$.  In the disjunctive model, in
the context where both arsonists drop a match, if the first arsonist
does not drop a match, the forest still burns down.  On the other hand,
in the conjunctive model, if the first arsonist does not drop a match (in
the same context), the forest does not burn down.

Note that the structural equations are deterministic.  
We later add probability to the picture by putting a probability on the set
of contexts (i.e., on the possible worlds).  This probability is not
needed in
the definition of causality, but will be useful in the
discussion of explanation.

\subsection{The Definition of Cause}\label{sec:actcaus}

With all this notation in hand, we can
now give a definition of actual cause (``cause'' for
short).  We want to make
sense out of statements of the form ``event $A$ is
an actual cause of
event $\phi$ (in context $\vec{u}$)''.%
\footnote{Note that we are using the word ``event'' here in the
standard sense of ``set of possible worlds''
(as opposed to ``transition between states of affairs''); 
essentially we are
identifying events with propositions.}
The picture here is that the context (and the structural equations) are
given.
Intuitively, they encode the background knowledge.
All the relevant 
events
are known.  The only question is picking
out which of them are the causes of $\phi$
or, alternatively, testing whether a given set of 
events
can be considered the cause of $\phi$.

The types of events that we 
allow
as actual causes are ones of
the form
$X_1 = x_1 \land \ldots \land X_k = x_k$---that is, conjunctions of
primitive events; we typically abbreviate this as $\vec{X} = \vec{x}$.
The events that can be caused are arbitrary Boolean combinations
of primitive events.
We do not believe that we lose much by disallowing disjunctive
causes here.   
Disjunctive {\em explanations}, however, are certainly of interest.

Roughly speaking, $\vec{X} = \vec{x}$ is a {\em sufficient cause of
$\phi$ in situation $(M,\vec{u})$\/} if 
(a) $\vec{X} = \vec{x} \land \phi$ is true in $(M, \vec{u})$ and (b) 
there is some set $\vec{x}'$
of values of $\vec{X}$ and some other variables $\vec{W}$ 
(satisfying some constraints) 
such that setting $\vec{W}$ to $\vec{w}'$ and
changing $\vec{X}$ to $\vec{x}'$ results in $\neg \phi$ being true
(that is,  $(M,\vec{u}) \sat [\vec{X} \gets \vec{x}', \vec{W} \gets
\vec{w}'] \neg \phi$).  
Part (b) is very close to the standard counterfactual
definition of causality, advocated by Lewis \citeyear{Lewis73} and
others: $\vec{X} = \vec{x}$ is the cause of $\phi$ if,
had $\vec{X}$ not been equal to $\vec{x}$, $\phi$ would not have been
the case.  The difference is that we allow setting some other variables
$\vec{W}$ to $\vec{w}$.  
As we said,
the formal definition 
puts some constraints on $\vec{W}$.  Among other things, it must be the
case that setting $\vec{X}$ to $\vec{x}$ is enough to force $\phi$ to be
true when $\vec{W}$ is set to $\vec{w}'$.  That is, it must be the case that
$(M,\vec{u}) \sat [\vec{X} \gets \vec{x}, \vec{W} \gets \vec{w}'] \phi$.
(See the appendix for the formal definition.)

$\vec{X} = \vec{x}$ is an {\em actual cause of $\phi$ in $(M,\vec{u})$\/} if
it is a sufficient clause with no irrelevant conjuncts.  That is,
$\vec{X} \gets \vec{x}$ is an actual cause of $\phi$ in $(M,\vec{u})$ if
it is a sufficient cause and no subset of $\vec{X}$ is also a sufficient
cause.   Eiter and Lukasiewicz \citeyear{EL01}
and, independently, Hopkins \citeyear{Hopkins01} have shown that actual
causes are always single conjuncts.  As we shall see, this is not the
case for explanations. 

Returning to Example~\ref{xam:arson}, note that $\ML_1 = 1$ is an actual
cause of $\FB=1$ in both the conjunctive and the disjunctive scenarios.
This should be clear in the conjunctive scenario.  Setting $\ML_1$ to 0
results in the forest not burning down.  To see that $\ML_1 = 1$ is also
a cause in the disjunctive scenario, let $W$ be $\ML_2$.  Note that
$(M_1, u_{00}) \sat [\ML_1 \gets 0, \ML_2 \gets 0] \FB = 0$, so the
counterfactual condition is satisfied.  Moreover, 
$(M_1, u_{00}) \sat [\ML_1 \gets 1, \ML_2 \gets 0] \FB = 1$; that is, in
the disjunctive scenario, if the first arsonist drops the match, that is
enough to burn down the forest, no matter what the second arsonist does.
In either scenario, both arsonists dropping a lit match 
constitutes a sufficient cause for the forest fire, as does the first
arsonist dropping a lit match and sneezing.  Given an actual cause, a
sufficient cause can be obtained by adding arbitrary conjuncts to it.

Although  each arsonist is a cause of the forest burning
down in the conjunctive scenario, under reasonable assumptions about the
knowledge of the agent wanting an explanation, each arsonist alone is not
an {\em explanation\/} of the forest burning down.  Both arsonists
together provide the explanation in that case;
identifying arsonist 1 would only trigger 
a
further quest for the identity of her accomplice.
In the disjunctive
scenario, each arsonist alone is an explanation of the forest burning down.

We hope that this informal discussion of the definition of causality
will suffice for 
readers who want to focus mainly on explanation.
For completeness, we give the formal definitions of sufficient and
actual cause  in the appendix.  
The definition is motivated, discussed, and defended in much more detail in 
Part I, where it is also compared with other definitions of causality.  
In particular, it is shown to avoid a number of problems
that have been identified with Lewis's account (e.g., see
\cite[Chapter 10]{pearl:2k}), such as commitment to transitivity of
causes.
For the purposes of this paper, we ask that the reader accept 
our definition of causality.
We note that, to some extent, our definition of explanation
is modular in its use of causality, in that another 
definition of causality could be substituted for the one we use in the
definition of explanation (provided it was given in the same framework).

\section{Explanation: The Basic Definition}\label{sec:explanation}
As we said in the introduction, many definitions of causal explanation
have been given in the literature. 
The ``classical'' approaches in the philosophy literature,
such as Hempel's \citeyear{Hempel65} {\em deductive-nomological\/}
model and 
Salmon's \citeyear{Salmon89} {\em statistical relevance\/} model (as
well as many other approaches), because they are based on 
logical implication and probabilistic dependence,
respectively, fail to exhibit the directionality inherent in common
explanations.  Despite all the examples in the philosophy literature on
the need for 
taking causality and counterfactuals into account, and the extensive work
on causality defined in terms of counterfactuals in 
the philosophy literature, as Woodward \citeyear{Woodward01} observes,
philosophers have been reluctant to build a theory of
explanation on top of a theory of  causality. The concern seems to be
one of circularity.
In this section, we give a definition of explanation based on the
definition of causality discussed in Section~\ref{sec:actcaus}.  
Circularity is avoided  because the definition of explanation invokes strictly 
formal features of a causal model that do not in turn depend on the
notion of explanation. 

Our definition of causality assumed that the causal model and all the
relevant facts were given; the problem was to figure out which of the
given facts were causes.
In contrast, the role of explanation is to provide the
information needed to establish causation.
Roughly speaking,
we view an explanation as a fact
that is not known for certain but, if found to be true,
would constitute a genuine
cause of the {\em explanandum\/} (fact to be explained), regardless of
the agent's initial uncertainty.  Roughly speaking, the role of
explanation is to provide the information needed to establish causation.
Thus, as we said in the introduction,
what counts as an explanation depends on what one already knows (or
believes---we largely blur the distinction between knowledge and belief
in this paper).  As a consequence, the definition of 
an {\em explanation\/} should be relative to the agent's
epistemic state (as in G\"ardenfors \citeyear{Gardenfors1}).
It is also  natural, from this viewpoint, that an
explanation will include fragments of the causal model $M$
or reference to the physical laws underlying the
connection between the cause and the effect.

The definition of explanation is motivated by the
following intuitions.  An individual in a given epistemic state $K$ asks
why $\phi$ holds.  What constitutes a good answer to his
question?  A good answer must (a) provide information that goes beyond
$K$ and (b) be such that the individual can see that it would, if true, be
(or be very likely to be) a cause of $\phi$.  We may also want to
require that (c) $\phi$ be true (or at least probable).  Although our
basic definition does not insist on (c), it is easy to add this requirement.

How do we capture an agent's epistemic state in our framework?  
For ease of exposition, we first consider
the case where the causal model is known
and only the context is uncertain.  (The minor
modifications required to deal with the general case
are described in Section~\ref{sec:general}.)  In that case, one way of
describing an agent's epistemic state 
is
by simply describing the set
of contexts the agent considers possible.

\dfn\label{def:explanation1} (Explanation)  Given a structural model
$M$, {\em $\vec{X} = \vec{x}$ is an explanation
of $\phi$ relative to a set $\K$ of contexts\/} if the following
conditions hold:
\begin{description}
\item[{\rm EX1.}] $(M,\vec{u}) \sat \phi$ for each context $\vec{u} \in \K$.
(That is, $\phi$ must hold in all contexts the agent considers
possible---the agent considers what she is trying to explain
as an established fact.)
\item[{\rm EX2.}] $\vec{X} = \vec{x}$ is a
{\em sufficient cause\/} of $\phi$ in $(M,\vec{u})$
for each $\vec{u} \in \K$ such that $(M,\vec{u}) \sat \vec{X} = \vec{x}$.
\item[{\rm EX3.}] $\vec{X}$ is minimal; no  subset of $\vec{X}$
satisfies EX2.
\item[{\rm EX4.}]
$(M,\vec{u}) \sat \neg (\vec{X}=\vec{x})$ for some $\vec{u} \in \K$ and
$(M,\vec{u}') \sat \vec{X}=\vec{x}$ for some $\vec{u}' \in \K$.
(This just says that the agent considers
a context possible where the explanation is false, so the explanation
is not known to start with, and considers a context possible where the
explanation is true, so that it is not vacuous.) \eprf
\end{description}
\end{definition}

Our requirement EX4 that the explanation is not known may seem
incompatible with linguistic usage.  Someone discovers some fact $A$ and
says ``Aha! That explains why $B$ happened.''  Clearly, $A$ is not
an explanation of why $B$ happened
relative to the epistemic state {\em after\/} $A$ has been discovered,
since at that point $A$ is known.  However, $A$ can legitimately be
considered an explanation of $B$ relative to the epistemic state before
$A$ was discovered.  
Interestingly, as we shall see, although there is a cause for every
event $\phi$ (although sometimes it may be the trivial cause $\phi$),
one of the effects of the requirement 
EX4 is that some events may have no explanation.  This seems to us quite
consistent with standard usage of the notion of explanation.  After all,
we do speak of ``inexplicable events''.

What does the definition of explanation tell us for the arsonist example?
What counts as a cause is, as expected, very much dependent on the
causal model and the agent's epistemic state.
If the causal model has only arsonists as the cause of the fire, there are two
possible explanations in the disjunctive scenario: (a) arsonist 1 
did it and (b) 
arsonist 2 did it (assuming $\K$ consists of three contexts, where
either 1, 2, or both set the fire).  In the conjunctive scenario, no
explanation is necessary, since the agent knows that both arsonists must
have lit a match if arson is the only possible cause of the fire
(assuming that the agent considers the two arsonists to be the only
possible arsonists).  
That is, if the two arsonists are the only possible cause of the fire
and the fire is observed, then $\K$ can consist of only one context,
namely, the one where both arsonists started the fire.  No explanation
can satisfy EX4 in this case.

Perhaps more interesting is to consider a causal model with other
possible causes, such as lightning and unattended campfires.
Since the agent knows that there was a fire, in each of the contexts in
$\K$, at least one of the potential causes must have actually occurred.
If there is a context in $\K$ where only arsonist 1 dropped a lit match
(and, say, there was lightning), another where only arsonist 2
dropped a lit match,
and a third where both arsonists dropped matches,
then, in the conjunctive scenario, $\ML_1 = 1 \land \ML_2 = 1$ 
is an
explanation of $\FB=1$, but neither $\ML_1=1$ nor $\ML_2=1$ by itself is
an
explanation (since neither by itself is a cause in all contexts in $\K$
that satisfy the formula).  On the other hand, in the disjunctive
scenario,
both $\ML_1 = 1$ and $\ML_2=1$ are explanations.

Consider the following example, due to Bennett (see
\cite[pp.~222--223]{sosa:too93}), which is analyzed in Part I.
\xam\label{xam1}   Suppose that there was a heavy rain
in April and electrical storms in the following two months; and in June
the lightning took hold.  If it hadn't been for the heavy rain in April,
the forest would have caught fire in May.  The first question is 
whether the April rains caused the forest fire.  According to  a
naive counterfactual analysis, they do,
since if it hadn't rained, there wouldn't have
been a forest fire in June.   In our framework,
it is not the case that the April rains caused
the fire, but they were a cause of there being a fire
in June, as opposed to May.  This seems to us intuitively right.

The situation can be captured using a model with three
endogenous random variables:
\begin{itemize}
\item $\AS$ for ``April showers'', with two
values---0 standing for did {\em not\/} rain heavily in April and 1
standing for rained heavily in April;
\item $\ES$ for ``electric storms'', with four possible values: $(0,0)$
(no electric storms in either May or June), (1,0) (electric storms in
May but not June), (0,1) (storms in June but not May), and (1,1)
(storms in both April and May);
\item and $F$ for ``fire'', with three possible values: 0 (no fire at
all), 1 (fire in May), or 2 (fire in June).
\end{itemize}
We do not describe the context explicitly.   Assume its value
$\vec{u}$ is such that it ensures that there is a shower in April, there
are electric storms in both May and June,
there is sufficient oxygen, there are no other potential causes of fire
(like dropped matches), no other inhibitors of fire (alert campers
setting up a bucket brigade), and so on.  That is, we choose $\vec{u}$
so as to allow us to focus on the issue at hand and to ensure that the
right things happened (there was both fire and rain).
We will not bother writing out the details of the structural
equations---they should be obvious, given the story (at least, for the
context $\vec{u}$); this is also the case for all the other examples in
this section. The causal network is simple: there are edges from
$\AS$ to  $F$ and from $\ES$ to $F$.  As observed in Part I,
each of the following holds.
\begin{itemize}
\item $\AS=1$ is a cause of the June fire $(F=2)$.
\item $\AS=1$ is {\em not\/} a cause of the fire ($F=1 \lor F=2$).  If
$\ES$ is set 
(0,1),  (1,0), or (1,1), then there will be a fire (in either May or
June) whether $\AS=0$ or $\AS=1$.  On the other hand, if $\ES$ is set to
(0,0), then there is no fire, whether $\AS=0$ or $\AS=1$.
\item $\ES = (1,1)$ is a cause of both $F=2$ and $(F=1 \lor F=2)$.
Having electric storms in both May and June caused there to be a fire.
\item $\AS=1 \land \ES=(1,1)$ is a sufficient cause of $F=2$; each
individual conjunct is an actual cause.
\end{itemize}

Now consider the problem of explanation.
Suppose that the agent knows that there was an electric
storm, but does not know when, and does not know whether there were
April showers.  Thus, $\K$ consists of six contexts,
one corresponding to each of the values
(1,0), (0,1), and (1,1) of $\ES$ and the values 0 and 1 of $\AS$.
Then it is easy to see that $\AS=1$ is not an explanation of fire ($F=1
\lor F=2$), since it is
not a cause of fire in any context in $\K$.  Similarly, $\AS=0$ is
not an explanation of fire.  On the other hand, each of $\ES=(1,1)$,
$\ES=(1,0)$, and $\ES=(0,1)$ is an explanation of fire.

Now suppose that we are looking for an explanation of the June fire.
Then the set $\K$ can consist only of contexts
compatible with there being a fire in June.  Suppose that $\K$ consists
of three contexts, one where $\AS=1$ and $\ES = (0,1)$, one where $\AS=1$
and $\ES=(1,1)$, and one where $\AS=0$ and $\ES=(0,1)$.  In this case, each
of $\AS=1$, $\ES=(0,1)$, and $\ES=(1,1)$ is
an explanation
of the June fire.
(In
the case of $\AS=1$, we need to consider the setting where $\ES=(1,1)$.)

Finally, if the agent knows that there was an electric storm in May and
June and heavy rain in April (so that $\K$ consists of only one
context), then there is no explanation of either fire or the fire in
June.  Formally, this is because it is impossible to satisfy EX4.
Informally, this is because the agent already knows why there was a
fire in June.
\exam

Note that, as for causes, we have disallowed disjunctive explanations.
Here the motivation is less clear cut.  It does make perfect sense to
say that the reason that $\phi$ happened is either $A$ or $B$ (but I
don't know which).  There are some technical difficulties with
disjunctive explanations, which suggest philosophical problems.  For
example,
consider the conjunctive scenario of the arsonist example again. Suppose
that the structural model is such that the only  causes of fire are the
arsonists, lightning, and unattended campfires and that $\K$ consists of
contexts where each of these possibilities is the actual cause of the
fire.  Once we allow disjunctive explanations, what is the explanation
of fire?  One candidate is ``either there were two arsonists or there
was lightning or there was an unattended campfire (which got out of
hand)''.  But this does not satisfy EX4, since the disjunction is true
in every context in $\K$.  
On the other hand, the disjunction of any two of the three clauses does
satisfy EX4.  As we add more and more potential causes of the fire
(explosions, spontaneous combustion, \ldots) larger and larger
disjunctions will 
count as explanations.  The only thing that is not allowed is the
disjunction of all possible causes.  
This distinction between the dijsunction of all potential causes and
the disjunction of all but one of the potential causes seems artificial,
to say the least.
To make matters worse,
there is the technical problem of reformulating the minimality condition
EX3 to deal with disjunctions.
We could not see any reasonable way to
deal with these technical problems, so we ban disjunctive
explanations.

We believe that, in cases where disjunctive explanations seem
appropriate, it is best to capture this directly in the causal model
by having a variable that represents the disjunction.
(Essentially the same point is made 
by Chajewska and Halpern \citeyear{CH97}.)  For
example, consider the disjunctive scenario of the arsonist example,
where
there are other potential causes of the fire.  If we want to allow
``there was an arsonist'' to be an explanation without specifically
mentioning who the arsonist is, then it can be easily accomplished by 
replacing the variables $\ML_1$ and $\ML_2$ in the model by a variable
$\ML$ which is 
1 iff at least one arsonist drops a match.
Then $\ML=1$ becomes an explanation, without requiring disjunctive
explanations.

\commentout{
Why not just add $\ML$ to the model rather than using it to replace
$\ML_1$ and $\ML_2$?  We have implicitly assumed in our framework that 
all possible combinations of assignments to the variables are possible
(i.e., there is a structural contingency for any setting of the
variables).  If we add $\ML$ and view it as being logically equivalent to
$\ML_1 \lor \ML_2$ (that is, $\ML=1$ {\em by definition\/} iff at least
one of $\ML_1$ and $\ML_2$ is 1) then, for example, it is logically
impossible for there to be a structural contingency where $\ML_1 = 0$,
$\ML_2 = 0$, and $\ML = 1$.  Thus, in the presence of logical
dependences, it seems that we need to restrict the set of contingencies
that can be considered to those that respect the dependencies.  We have
not yet considered the implications of such a change for our framework,
so we do not pursue the matter here.%
\footnote{As a related
matter, note a model with logical dependencies means the model is no longer
recursive.}
}

One other point regarding the definition of explanation:
When we ask for an explanation of $\phi$, we usually expect that it not
only explains $\phi$, but that it is true in the actual world.
Definition~\ref{def:explanation1} makes no reference to the ``actual
world'', only to the agent's epistemic state.  There is no difficulty
adding an actual world to the picture and requiring that the explanation
be true in the world.  We can simply talk about an explanation relative
to a pair $(\K,\vec{u})$, where $\K$ is a set of contexts.
Intuitively, $\vec{u}$ describes the context in the actual world.  The
requirement that the explanation be true in the actual world then 
becomes $(M,\vec{u}) \sat \vec{X} = \vec{x}$.  Although we have not made
this requirement part of the definition, adding it would 
have no significant effect on our discussion.
Once we have the actual world as part of the model, we could also 
require that $\vec{u} \in \K$.  This condition would entail that $\K$ 
represents the agent's knowledge rather than the agent's beliefs: the
actual context is one of the ones the agent considers possible.  

\section{Partial Explanations and Explanatory Power}\label{s:partialexp}
Not all explanations are considered equally good.  Some
explanations are more likely than others.  
One way to
define the ``goodness'' of an explanation is by bringing
probability into the picture.  Suppose that the
agent has a probability on the set $\K$ of possible contexts.  In this
case, we can consider the probability of the set of contexts where the
explanation $\vec{X} = \vec{x}$ is true.   For example, if the agent has
reason to believe that electric storms are quite common in both May and
June, then the set of contexts where $\ES=(1,1)$ holds would have greater
probability than the set where either $\ES=(1,0)$ or $\ES=(0,1)$ holds.
Thus, $\ES=(1,1)$ would be considered a better explanation.

Formally, suppose that there is a probability $\Pr$ on the set $\K$ of
possible contexts.  Then the probability of explanation $\vec{X} = \vec{x}$
is just $\Pr(\vec{X} = \vec{x})$.  While the probability of an
explanation clearly captures some important aspects of how good the
explanation is, it is only part of the story.
The other part concerns the degree to which an explanation
fulfills its role (relative to $\phi$) in the various contexts
considered.
This becomes clearer when we consider {\em partial\/} explanations.
The following example, taken from \cite{Gardenfors1},
is one where partial explanations play a role.

\xam\label{xam7}  Suppose I see that Victoria is tanned and I seek an
explanation.  Suppose that the causal model includes variables for
``Victoria took a vacation in the Canary Islands'', ``sunny in the
Canary
Islands'',  and ``went to a tanning salon''.
The set $\K$ includes
contexts for all settings of these variables compatible with Victoria
being tanned.
Note that, in particular, there is a context where Victoria went both
to the Canaries (and didn't get tanned there, since it wasn't sunny)
and to a tanning salon. G\"ardenfors points out that we normally accept
``Victoria took a vacation in the Canary Islands'' as a satisfactory
explanation of Victoria being tanned and, indeed, according to his
definition, it is an explanation.
Victoria taking a vacation is not an
explanation (relative to the context $\K$) in our framework,
since
there is a context $\vec{u}^* \in \K$ where Victoria went to the Canary
Islands but it was not sunny,
and in $\vec{u}^*$ the actual cause of her tan is
the tanning salon, not the vacation.
Thus, EX2 is not satisfied.  However, intuitively, it is ``almost''
satisfied, since it is satisfied by every context in $\K$  in which
Victoria goes to the Canaries but $u^*$.
The only complete explanation according to our definition is
``Victoria went to the Canary Islands {\em and\/} it was sunny.''
``Victoria went to the Canary Islands'' is  a {\em partial\/}
explanation, in a sense to be defined below.
\exam

In Example~\ref{xam7}, the partial explanation can be extended to a complete
explanation by adding a conjunct.
But not every partial explanation can be extended
to a complete explanation.
Roughly speaking, the complete explanation may involve exogenous factors,
which are not permitted in explanations.
 Assume, for example, that going to a tanning
salon was not an endogenous variable in the model;
instead, the model simply had an exogenous variable $U_s$ that could make
Victoria suntanned even in the absence of sun
in the Canary Islands. Likewise, assume that the weather in
the Canary Islands was also part of the background context.
In this case,  
Victoria's vacation would still be a 
partial explanation of her suntan, since the context where
it fails  to be a cause (no sun in the Canary Islands)
is fairly unlikely, but we cannot add conjuncts to this
event to totally exclude that context from the
agent's realm of possibilities.
Indeed, in this model there is no (complete) explanation for Victoria's
tan; it is inexplicable!
Inexplicable events are not so uncommon, as the following example shows.

\begin{figure}[htb]
\input{psfig}
\centerline{\psfig{figure=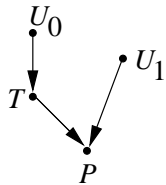}}
\caption{The television with no picture.}
\label{fig-z}
\end{figure}

\xam\label{xam8} Suppose that the sound on a television works but there
is
no picture.  Furthermore, the only cause of there being no picture that
the agent is aware of
is the picture tube being faulty.  However, the agent is also
aware that there are times when there is no picture even though the
picture tube works perfectly well---intuitively, there is no picture
``for inexplicable 
reasons''.  This is captured by the causal network described in Figure~\ref{fig-z}, where
$T$ describes whether or not the picture tube is working (1 if it is
and 0 if it is not) and $P$
describes whether or not there is a picture (1 if there is and 0 if
there is not).
The exogenous variable $U_0$ determines the status of the picture tube:
$T = U_0$.
The exogenous variable
$U_1$ is meant to represent the mysterious ``other possible causes''.
If $U_1=0$, then whether or not there is a picture depends solely on the
status of the picture tube---that is, $P = T$.
On the other hand, if
$U_1 = 1$, then there is no picture ($P=0$) no matter what the status of
the
picture tube.  Thus, in contexts where $U_1=1$, $T=0$ is {\em not\/} a
cause of $P=0$.  Now suppose that $\K$ includes a context
$\vec{u}_{00}$ where $U_0=U_1 =0$
and $\vec{u}_{10}$ where $U_0=1$ and $U_1 =0$.
The only cause of $P=0$ in both $\vec{u}_{00}$ and $\vec{u}_{10}$ is $P=0$
itself. 
($T=0$ is not a cause of $P=0$ in $\vec{u}_{10}$, since $P=0$ even if $T$ is
set to 1.)
As a result, there is no explanation of $P=0$ relative to an
epistemic $\K$ that includes $\vec{u}_{00}$ and $\vec{u}_{10}$.
(EX4 excludes the vacuous explanation $P=0$.)
On the other hand, $T=0$ is a cause of $P=0$ in all other contexts in
$\K$ satisfying $T=0$ other than $\vec{u}_{00}$.   If the probability of
$\vec{u}_{00}$  
(capturing the intuition that it is unlikely that more than one thing
goes wrong with a television at once), 
then we are entitled to view
$T=0$ as a quite good partial explanation of $P=0$ with respect to $\K$.

Note that if we modify the causal model here by adding an
endogenous variable, say $I$, corresponding to the ``inexplicable'' cause
$U_1$ (with equation $I = U_1$), then $I=0$ is a cause of
$P=0$ in the context $u_{10}$ (and both $I=0$ and $T=0$ are causes of
$P=0$ in $u_{00}$).  In this model, $I = 0$ is an explanation of $P=0$.
\exam

Example~\ref{xam8} and the discussion after Example~\ref{xam7}
illustrate an important point.  Adding an endogenous variable
corresponding to an exogenous variable can result in there being an
explanation when there was none before.  This phenomenon of adding
``names'' to create explanations is quite common.  For example, ``gods''
are introduced to explain otherwise inexplicable phenomena; clusters of
symptoms are given names in medicine so as to serve as explanations.

In any case, these examples motivate the following definition of partial
explanation. 
\dfn\label{def:partial} 
Let $\K_{\vec{X} = \vec{x},\phi}$ be the largest subset $\K'$ of $\K$
such 
that $\vec{X} = \vec{x}$ is an explanation of $\phi$ relative to
$\K_{\vec{X} = \vec{x},\phi}$.
(It is easy to see that there is a largest such set; 
it consists of all the contexts in $\K$ except the ones where
$\vec{X}=\vec{x}$ is true but is not a sufficient cause of $\phi$.)
Then $\vec{X} = \vec{x}$ is a {\em partial explanation of $\phi$ with
goodness $\Pr(\K_{\vec{X}= \vec{x},\phi}|\vec{X} = \vec{x})$}.
Thus, the goodness of a partial explanation measures the extent to which
it provides an explanation of $\phi$.%
\footnote{Here and elsewhere, a formula such as $\vec{X} = \vec{x}$ is
being identified with the set of contexts where the formula is true.
Recall that, since all contexts in $\K$ are presumed
to satisfy $\phi$, there is no need to condition
on $\phi$; this probability is
already updated with the truth of the explanandum $\phi$.
Finally, note that our usage of partial explanation is related to,
but different from, that of Chajewska and Halpern \citeyear{CH97}.}
\edfn

In Example~\ref{xam7}, if the agent believes that it is sunny in the
Canary Islands with probability .9
(that is, the probability that it was sunny given that Victoria is
suntanned and that she went to
the Canaries is .9),
then Victoria going to the Canaries is a
partial explanation of her being tanned with goodness .9.  The 
relevant set $\K'$
consists of those contexts where it is sunny in the Canaries.
Similarly, in Example~\ref{xam8}, if the agent believes that the
probability of both the picture tube being faulty and the other
mysterious causes being operative is .1, then $T=0$ is a partial
explanation of $P=0$ with goodness .9 (with $\K'$ consisting of all the
contexts where $U_1=1$).

A full explanation is clearly a partial explanation with goodness 1, but
we are 
often satisfied with partial 
explanations $\vec{X}=\vec{x}$ that are not as good,
especially if they have high probability 
(\ie if $\Pr(\vec{X} = \vec{x})$ is high).
Note that,
in general, there 
is a
tension
between the goodness of an explanation and its probability.

These ideas also lead to a definition of {\em explanatory power}.
Consider Example~\ref{xam:arson} yet again, and suppose that there is an
endogenous random variable $O$ corresponding to the presence of oxygen.
Now if $O=1$ holds in all the contexts that the agent considers
possible, then $O=1$ is excluded as an explanation by EX4.  
If the agent
knows that there is oxygen, then the presence of oxygen cannot be part
of an explanation. 
But suppose that $O=0$ holds in one context that the agent considers
possible, albeit a very unlikely one (for example, there may be another
combustible gas).  In that case, $O=1$ becomes a very good partial
explanation of the 
fire.  Nevertheless, it is an explanation with, intuitively, very little
explanatory power.  How can we make this precise?

Suppose that there is a probability distribution $\Pr^-$ on a set $\K^-$
of contexts that includes $\K$.  $\Pr^-$ intuitively represents the
agent's ``pre-observation'' probability, that is, the agent's
prior probability before the explanandum $\phi$ is observed or discovered.
Thus, $\Pr$ 
is the result of conditioning $\Pr^-$ on $\phi$
and $\K$ consists of the contexts in $\K^-$ that satisfy $\phi$.  
 G\"ardenfors identifies the explanatory
power of the (partial) explanation $\vec{X} = \vec{x}$ of $\phi$ 
with $\Pr^-(\phi|\vec{X} = \vec{x})$  (see \cite{CH97,Gardenfors1}).
If this probability is higher than $\Pr^-(\phi)$,
then the explanation makes $\phi$ more likely.%
\footnote{Actually, for G\"ardenfors, the explanatory power of $\vec{X}
= \vec{x}$ is $\Pr^-(\phi|\vec{X} = \vec{x}) - \Pr^-(\phi)$.  But as far
as comparing the explanatory power of two explanations, it suffices to
consider just $\Pr^-(\phi|\vec{X} = \vec{x})$, since the $\Pr^-(\phi)$
terms will appear in both expressions.  We remark that Chajewska and
Halpern \citeyear{CH97} argued that the quotient
$\Pr^-(\phi|\vec{X})/\Pr^-(\phi)$ gives a better measure of explanatory
power than the difference, but the issues raised by Chajewska and
Halpern are irrelevant to our concerns here.}
Note that since $\K$ consists of all the contexts in $\K^-$ where $\phi$
is true, G\"ardenfors' notion of explanatory power is equivalent to
$\Pr^-(\K| \vec{X} = \vec{x})$.

G\"ardenfors' definition clearly captures some important features of our
intuition.  For example, under reasonable assumptions about
$\Pr^-$, $O=1$ has much lower explanatory power than, say, $\ML_1=1$.
Learning that there is oxygen in the air certainly has almost no
effect on an agent's prior probability that the forest burns down, while
learning that an arsonist dropped a match almost certainly increases it.
However, G\"ardenfors' definition still has a problem.  It basically
confounds correlation with causation.  
For example, according to this definition, the barometer
falling is an explanation of it raining with high explanatory power.

We would argue that a better measure of the explanatory power of
$\vec{X} = \vec{x}$ is $\Pr^-(\K_{\vec{X} = \vec{x},\phi}|\vec{X} =
\vec{x})$.   Note that the two definitions agree in the case that
$\vec{X} = \vec{x}$ 
is a full explanation (since then $\K_{\vec{X} = \vec{x},\phi}$ is just
$\K$, the set of contexts in $\K^-$ where $\phi$ is true).
In particular, they agree that $O=1$ has very low explanatory power,
while $\ML_1 = 1$ has high explanatory power.  
The difference between the two definitions arises if there are contexts
where $\phi$ and $\vec{X} = \vec{x}$ both happen to be true, but
$\vec{X} = \vec{x}$ is not a cause of $\phi$.
In Example~\ref{xam7}, the context $\vec{u}^*$ is one such context,
since in $\vec{u}^*$, Victoria went to the 
Canary Islands, but this was not an explanation of
her getting tanned, since it was not sunny.  
Because of this difference, for us, the falling barometer has 0
explanatory power as far as explaining the rain.  Even though
the barometer falls in almost all contexts where it rains
(assume that there are contexts where it rains and the barometer does
not fall, perhaps because it is defective, so that the barometer falling
at least satisfies EX4), the barometer falling is not a {\em cause\/} of
the rain in any context.  Making the barometer rise would not result in
the rain stopping!

Again, (partial) explanations with higher explanatory power typically
are 
more refined and, hence, less
likely, than explanations with less explanatory power.
There is no obvious way to resolve this tension.
(See \cite{CH97} for more discussion of this issue.)

As this discussion suggests,
our definition shares some features with that of G\"ardenfors'
\citeyear{Gardenfors1}.  Like him, we consider explanation relative to
an agent's epistemic state.  
G\"{a}rdenfors also considers a 
``contracted'' epistemic state characterized by the distribution 
$\Pr^-$.  Intuitively, $\Pr^-$ describes the agent's beliefs before
discovering $\phi$.  (More accurately, it describes an epistemic state
as close as possible to $\Pr$ where the agent does not ascribe
probability 1 to $\phi$.)  
If the agent's current epistemic state came about as a result of
observing $\phi$, then we can take $\Pr$ to be the result of
conditioning $\Pr^-$ on $\phi$.
However, G\"{a}rdenfors does necessarily assume such a
connection between $\Pr$ and $\Pr^-$.  In any case, for G\"{a}rdenfors,
$\vec{X} = \vec{x}$ is an explanation of $\phi$ relative
to $\Pr$ if (1) $\Pr(\phi) = 1$, (2) $0 < \Pr(\vec{X} = \vec{x}) < 1$,
and (3) $\Pr^-(\phi|\vec{X} = \vec{x}) > \Pr^-(\phi)$.  
(1) is the probabilistic analogue of EX1.
Clearly, (2) is the probabilistic analogue of EX4.  Finally, (3) says
that learning the explanation increases the likelihood of $\phi$.
G\"{a}rdenfors focuses on 
the explanatory power of an
explanation, but does not take into account its prior probability.
As pointed out by Chajewska and Halpern \citeyear{CH97}, G\"ardenfors'
definition suffers from 
another defect: 
If there is an explanation of $\vec{Y} = \vec{y}$ at all, then
for all events $\vec{X} = \vec{x}$ 
such that $0 < \Pr(\vec{X}=\vec{x} \land \vec{Y}=\vec{y}) < 1$, $\vec{X} =
\vec{x} \land \vec{Y} = \vec{y}$ is an explanation  of $\vec{Y} = \vec{y}$.
Moreover, it has the highest possible explanatory power.
(Note that, in our definition, EX3 blocks $\vec{X} =
\vec{x} \land \vec{Y} = \vec{y}$ from being a cause of $\vec{Y} = \vec{y}$.)

In contrast to G\"ardenfors' definition, the dominant approach to
explanation in the AI literature, the {\em maximum a posteriori (MAP)\/}
approach (see, for example, \cite{DH90,Pearl,Shim91a}),
focuses on  the 
probability of the explanation,  
that is, what we have denoted $\Pr(\vec{X} = \vec{x})$.%
\footnote{The MAP literature typically considers $\Pr^-(\vec{X} = \vec{x} |
\phi)$, but this is equivalent to $\Pr(\vec{X} = \vec{x})$.}
The MAP approach is based on the intuition that
the best explanation for an observation is the state of the world (in
our setting, the context) that is most probable 
given the evidence.  
There are various problems with this approach (see \cite{CH97} for a
critique).  Most of them can be dealt with, except for the main one:
it simply ignores the issue of explanatory power.
%
%
An explanation like $O=1$ has a very high probability, even though it is
intuitively irrelevant to the forest burning down.
To remedy this problem, more intricate combinations of the 
quantities $\Pr(\vec{X} = \vec{x})$, $\Pr^-(\phi|\vec{X}=\vec{x})$, and 
$\Pr^-(\phi)$ have been suggested 
to quantify the causal relevance of $\vec{X} = \vec{x}$ on $\phi$ but,
as argued 
by Pearl \citeyear[p.~221]{pearl:2k}, without taking causality into
account, no such 
combination of parameters can work.

\section{The General Definition}\label{sec:general}
In general, an agent may be uncertain about the causal model, so
an explanation will have to include information about it.
(G\"ardenfors \citeyear{Gardenfors1} and Hempel
\citeyear{Hempel65}
make similar observations, although they focus not on causal
information, but on statistical and nomological information;
we return to this
point below.)  It is relatively straightforward to extend our definition
of explanation to accommodate this provision.
Now an epistemic state $\K$ consists not
only of contexts, but of pairs $(M,\vec{u})$ consisting of a causal
model $M$ and a context $\vec{u}$.  
(Recall that such a pair is a {\em
situation}.)  Intuitively, now an explanation should consist of some
causal information (such as
``prayers do not cause fires'')
and the facts that are true.
Thus, a {\em (general) explanation\/} has the form $(\psi,
\vec{X}=\vec{x})$, where
$\psi$ is an arbitrary formula in our causal language and, as before,
$\vec{X}=\vec{x}$ is a conjunction of primitive events.
We think of the $\psi$ component as consisting of some causal information
(such as ``prayers do not cause fires'', which corresponds to a
conjunction of statements of the form 
$(F = i) \rimp [P \gets x](F=i)$, where $P$
is a random variable describing whether or not prayer takes place).
The first component in a general explanation
is viewed as restricting the set of causal models. To make this precise,
given a causal model $M$, we say $\psi$ is {\em valid in $M$\/}, and
write $M \sat \psi$, if $(M,\vec{u}) \sat \psi$ for all contexts
$\vec{u}$
consistent with $M$.
With this background, it is easy to state the general definition.

\dfn\label{def:explanation2} {\em $(\psi,\vec{X} = \vec{x})$ is an
explanation of $\phi$ relative to a set $\K$ of situations\/} if the
following conditions hold:
\begin{description}
\item[{\rm EX1.}] $(M,\vec{u}) \sat \phi$ for each situation $(M,\vec{u})
\in \K$.
\item[{\rm EX2.}]  $\vec{X} = \vec{x}$ is a sufficient cause of $\phi$ in
$(M,\vec{u})$ for all $(M,\vec{u}) \in \K$ such that
$(M,\vec{u}) \sat \vec{X} = \vec{x}$ and
$M \sat \psi$.
$\vec{X} = \vec{x}$ is a sufficient cause of $\phi$ in $(M,\vec{u})$.
\item[{\rm EX3.}]
$(\psi,\vec{X}=\vec{x})$ is minimal; there is no pair
$(\psi',\vec{X}' = \vec{x}') \ne (\psi,\vec{X} = \vec{x})$ satisfying EX2
such that $\{M'' \in M(\K): M'' \sat \psi'\} \supseteq
\{M'' \in M(\K): M'' \sat \psi\}$,
 where $\M(\K) = \{M: (M,\vec{u}) \in \K \mbox{ for some } \vec{u}\}$,
$\vec{X}' \subseteq \vec{X}$, and $\vec{x}'$ is the restriction of
$\vec{x}$ to the variables in $\vec{X}'$.
Roughly speaking, this says that no subset of $X$ provides a 
sufficient
cause of $\phi$ in more contexts than those where $\psi$ is valid.
\item[{\rm EX4.}]
$(M,\vec{u}) \sat \neg (\vec{X}=\vec{x})$ for some $(M,\vec{u}) \in \K$
and $(M',\vec{u}') \sat \vec{X}=\vec{x}$ for some $(M',\vec{u}') \in
\K$. \eprf
\end{description}
\end{definition}
Note that, in EX2, we now restrict the requirment of
sufficiency to situations $(M,\vec{u}) \in \K$
that satisfy both parts of the explanation $(\psi,\vec{X} = \vec{x})$, in
that $M \sat \psi$ and $(M,\vec{u}) \sat \vec{X} = \vec{x}$.
Furthermore, although both components of an explanation are
formulas in our causal language, they play very different roles.
The first component serves to restrict the set of causal models
considered (to those with the appropriate structure); the second
describes a cause of $\phi$ in the resulting set of situations.

Clearly Definition~\ref{def:explanation1} is the special case of
Definition~\ref{def:explanation2}
where there is no uncertainty about the causal structure
(i.e., there is some $M$ such that if $(M', \vec{u}) \in \K$, then $M =
M'$).  In this case, it is clear that we can take
$\psi$ in the explanation to be $\true$.

\xam
Using this general definition of causality, let us consider
Scriven's \citeyear{Scriven59} famous paresis example, 
which has caused problems for many other formalisms.
Paresis develops only in patients who
have been syphilitic for a long time, but only a small number of
patients who are syphilitic in fact develop paresis.  Furthermore, 
according to Scriven,
no other factor is known to be relevant in the development of paresis.%
\footnote{Apparently there are now other known factors, but this does
not change the import of the example.}
This description is captured by a simple causal model $M_P$.
There are two endogenous variables,
$S$ (for syphilis) and $P$ (for paresis), and two exogenous
variables, $U_1$, the background factors that determine $S$, and
$U_2$, which intuitively represents ``disposition to paresis'',
that is, the factors that determine, in conjunction with
syphilis, whether paresis actually develops.
An agent who knows this causal model and that a patient
has paresis does not need an explanation of why: he knows without
being told that the patient must have syphilis and that $U_2=1$.
On the other hand, for an agent who does not know the
causal model (i.e., considers a number of causal
models of paresis possible), 
$(\psi_P,S=1)$ is an explanation of paresis, where $\psi_P$ is a formula
that characterizes $M_P$.  
\exam

Definition~\ref{def:explanation2} can also be extended to deal naturally
with probability.  Actually, probability plays a role in two places.
First, there is  a probability on the situations in $\K$, analogous to
the probability on contexts discussed in Section~\ref{s:partialexp}.
Using this probability, it is possible to talk about the goodness of a
partial explanation and  to talk about explanatory power, just as
before.   In addition, probability can also be added to causal models to
get {\em probabilistic causal models}.  A probabilistic 
causal model is a tuple $M = (\S,\F,\Pr)$, 
where $M =(\S,\F)$ is a causal model and $\Pr$ is a probability measure
on the contexts defined by signature $\S$ of $M$.  In a probabilistic
causal model, it is possible to talk about the probability that
$X=3$; this is just the probability of the set of context in $M$ where
$X=3$.  More importantly for our purposes, it is also possible to say
``with probability .9, $X=3$ causes $Y=1$''.  Again, this is 
just the probability of the set of contexts where $X=3$ is the cause of
$Y=1$.  

Once we allow probabilistic causal models, explanations can include
statements like ``with probability .9, working with asbestos causes lung
cancer''.  Thus, probabilistic causal models can capture 
statistical information of the kind considered by G\"{a}rdenfors and
Hempel.   To fit such statements into the framework discussed above, we
must extend the language to allow statements about probability.  For example,
$\Pr([X\gets 3](Y=1)) = .9$, ``the probability that setting $X$ to 3
results in $Y$ being 1'', would be a formula in the language.  Such
formulas could then become part of the first component $\psi$ in a
general explanation.%
\footnote{We remark that G\"{a}rdenfors also consider two types of
probability measures, one on his analogue of situations, and one on the
worlds in a model.  Like here, the probability measure on worlds in a
model allows explanations with statistical information, while the
probability on situations allows him to define his notion of explanatory
power.}

\section{Discussion}\label{sec:discussion}
We have given a formal definition of explanation in terms of
causality.  As we mentioned earlier, there are not too many formal
definitions of explanation in terms of causality in the literature.
One of the few exceptions
is given by Lewis~\citeyear{Lewis86b}, who
defends the thesis that ``to explain an event is to provide some
information about its causal history''.  While this view is compatible
with our definition, there is no formal definition given to allow for a
careful comparison between the approaches.  
In any case, if we were to define causal history in terms of Lewis's
\citeyear{Lewis73} definition of causality, we would
inherit all the problems of that definition.
Our definition avoids these problems.

We have mentioned one significant problem of the definition
already: dealing with disjunctive explanations.  Disjunctions cause
problems in the definition of causality, which is why we do not deal
with them in the context of explanation.  As we pointed out
earlier, it may be possible to modify the definition of causality so as
to be able to deal with 
disjunctions
without changing the structure of
our definition of explanation.  
In addition, our definition gives no tools for dealing with the
inherent tension between explanatory power,
goodness of partial beliefs, and the probability of the explanation.
Clearly this is an area that requires further work.


\appendix
\section{Appendix: The Formal Definition of Causality}
To keep this part of the paper self-contained,
we reproduce here the formal definition of actual causality from
Part I.

\dfn\label{actcaus}
(Actual cause)
$\vec{X} = \vec{x}$ is an {\em actual cause of $\phi$ in
$(M, \vec{u})$ \/} if the following
three conditions hold:
\newcounter{enumerate}
\begin{description}
\item[{\rm AC1.}]\label{ac1} $(M,\vec{u}) \sat (\vec{X} = \vec{x}) \land
\phi$.
(That is, both $\vec{X} = \vec{x}$ and $\phi$ are true in the actual
world.)
\item[{\rm AC2.}]\label{ac2}
There exists a partition $(\vec{Z},\vec{W})$ of $\V$ with $\vec{X}
\subseteq \vec{Z}$ and some
setting $(\vec{x}',\vec{w}')$ of the variables in $(\vec{X},\vec{W})$
such that if $(M,\vec{u}) \sat (Z = z^*)$ for 
all
$Z \in \vec{Z}$, then
\begin{description}
\item[{\rm (a)}]
$(M,\vec{u}) \sat [\vec{X} \gets \vec{x}',
\vec{W} \gets \vec{w}']\neg \phi$.
In words, changing $(\vec{X},\vec{W})$ from $(\vec{x},\vec{w})$ to
$(\vec{x}',\vec{w}')$ changes
$\phi$ from true to false;
\item[{\rm (b)}]
$(M,\vec{u}) \sat [\vec{X} \gets
\vec{x}, \vec{W'} \gets \vec{w}', \vec{Z}' \gets \vec{z}^*]\phi$ for 
all subsets $\vec{W'}$ of $\vec{W}$ and all subsets $\vec{Z'}$ of $\vec{Z}$. 
In words, setting any subset of variables in $\vec{W}$ to their values\
in $\vec{w'}$ should have no 
effect on $\phi$ 
as long as $\vec{X}$ is kept at its current value $\vec{x}$,
even if all the variables in an arbitrary subset of $\vec{Z}$ are set to
their original values in the context $\vec{u}$.
\end{description}
\item[{\rm AC3.}] \label{ac3}
$\vec{X}$ is minimal; no subset of $\vec{X}$ satisfies
conditions AC1 and AC2.
Minimality ensures that only those elements of
the conjunction $\vec{X}=\vec{x}$ that are essential for
changing $\phi$ in AC2(a) are
considered part of a cause; inessential elements
are pruned. \bbox
\label{def3.1}  
\end{description}
\end{definition}

$\vec{X} = \vec{x}$ is a {\em sufficient cause\/} of $\phi$ in
$(M,\vec{u})$ if AC1 and AC2 hold, but not necessarily AC3.

We remark that in Part I, a slight generalization of this
definition is considered.  Rather than allowing {\em all\/} settings of
$\vec{X}$ and $\vec{W}$ in AC2, the more general definition presupposes
a set of allowable settings.  All the settings used in AC2 must come
from this allowable set.  It is shown that by generalizing in this way,
it is possible to avoid a number of problematic examples.

\subsection*{Acknowledgments:} Thanks to Riccardo Pucella and Vicky
Weissman for useful comments.

\bibliographystyle{chicago}
\bibliography{z,joe,refs,expl} 
\end{document}